\documentclass[conference]{IEEEtran}
\IEEEoverridecommandlockouts
\usepackage{cite}
\usepackage{amsmath,amssymb,amsfonts}
\usepackage{algorithmic}
\usepackage{graphicx}
\usepackage{textcomp}
\usepackage{xcolor}
\usepackage{booktabs}
\usepackage{multirow}
\def\BibTeX{{\rm B\kern-.05em{\sc i\kern-.025em b}\kern-.08em
    T\kern-.1667em\lower.7ex\hbox{E}\kern-.125emX}}
\usepackage{makecell}   
\usepackage{url}

\begin{document}

\title{
Controlling Implicit Shortcut Reliance in L2 Spoken English Auto-markers}

\author{
\IEEEauthorblockN{
Shilin Gao,
Mark J. F. Gales,
Kate M. Knill
}
\IEEEauthorblockA{
\textit{ALTA Institute, Machine Intelligence Laboratory, Cambridge University Engineering Department}  \\
Cambridge, UK \\
\texttt{\{sg2196,mjfg100,kmk1001\}@cam.ac.uk}
}
}

\maketitle

\begin{abstract}
Increasingly, speech and language processing tasks take either audio or text directly rather than extracting features from these as the input to the classifier or regressor. Often these systems make use of complex, for example transformer-based, processes that have the ability to derive highly non-linear mappings between the input and the output. Unfortunately these systems can also learn ``shortcuts'' where the classifier is overly reliant on particular aspects of the input to yield the output.
For the task of language proficiency assessment, this over-reliance can enable learners to increase their score by exploiting the shortcut rather than improving their ability. 
This paper introduces a novel training criterion that is able to reduce the classifier's reliance on shortcuts, thus for example limiting this option for malpractice in language assessment. This process is illustrated on two forms of assessment system, one based on the audio the other on the speech recognition text. 
The results show that, for both systems, there is higher correlations with features that could be exploited for malpractice than expected from the human reference, indicating an over-reliance on these features.
By introducing the modified training criterion, this correlation can be reduced to be  closer to the reference correlation.

\end{abstract}




\section{Introduction}

There is a growing adoption of end-to-end neural systems in speech and language processing, where raw audio or text is taken directly as input, rather than first extracting human understandable features. Systems based on models such as BERT~\cite{devlin2019}, ModernBERT~\cite{warner2024modernbert} and wav2vec~2.0.0~\cite{baevski2020} make use of complex, transformer-based processes capable of deriving highly non-linear mappings between input and output. Whilst they have brought substantial performance gains across a wide range of tasks, they are also more vulnerable to the risk of \textit{shortcut learning}~\cite{hermann2024foundations}. Here the model makes uses of aspects (features) of the data not only based on how reliably that feature relates to training-set labels, but also on how easily the feature can be extracted from inputs, leading to over-reliance on surface aspects to produce the output.

For the task of language proficiency assessment, this over-reliance has direct practical consequences: in a study designed to ``trick'' an automated essay scorer, the most effective submission simply repeated the same paragraphs 37 times, and got the highest possible score from the system while receiving the lowest possible score from the human reader~\cite{powers2002stumping}.
For spoken language, the analogue is straightforward: a learner who simply keeps talking, even repetitively, could exploit the same length-correlated shortcut.
The concern that an auto-marker can be exploited by test-takers to inflate scores without improving their underlying ability is well-established in Automated Essay Scoring~(AES).
It has long been shown that automated graders can over-rely on essay length~\cite{chodorow2004beyond,attali2007construct,fleckenstein2020long}, and that a model exploiting length alone can achieve state-of-the-art performance on benchmark datasets~\cite{jeon-strube-2021-countering}.
This has given rise to studies aiming at improving robustness around grader validity, including rationale alignment via counterfactuals~\cite{wang-etal-2024-beyond-agreement}, 
detection of adversarially designed inputs~\cite{farag-etal-2018-neural}, 
and human-in-the-loop supplementation of automated predictions~\cite{chakravarty2025enhancing}.
In Spoken Language Assessment (SLA), temporal fluency features such as speech rate, pause distribution and word count have long been established as predictors of oral proficiency~\cite{wang2018towards, bamdev2023automated, handley2024measures}, and critically examined for their construct validity~\cite{handley2024measures} but are at risk of being capitalised on.


Since these features both correlate with human judgements and can be exploited if the auto-marker puts too much reliance on them, the goal is to mitigate the latter whilst balancing the contribution of a feature to the predicted score.
Existing mitigation strategies, such as feature reweighting and length normalisation~\cite{chodorow2004beyond, attali2007construct}, gradient-based penalties~\cite{ross2017right, shao2021right}, and attribution-based defences~\cite{sundararajan2017axiomatic, kumar-etal-2023-automatic, wang-etal-2024-beyond-agreement}, all require the shortcut feature to be an explicit, differentiable input to the model.
Fine-tuned encoder-based graders, which take raw audio or speech transcripts directly, do not satisfy this condition:
e.g. response length is absorbed implicitly into learned representations with no direct gradient path from the predicted score to this proxy feature.
We propose a training objective that penalises over-dependence on any externally computable proxy feature by adding a rank correlation penalty term to the primary task loss.
Both losses are rank-based, ensuring comparability across features with differing dynamic ranges.
The penalty operates purely at the output level without modifying the encoder or requiring explicit feature access, and is applicable to both text-based and audio-based graders.

We demonstrate the method on the Speak~\&~Improve 2025 Corpus~\cite{sicorpus25,qian2025speak} across four task types, using a ModernBERT text grader~\cite{raina20_interspeech,qian25_slate} and a wav2vec~2.0 audio grader~\cite{banno2023wav2vec}, with word count and Voice Activity Detection (VAD) time as the respective shortcut proxies. Zero-shot Qwen2.5-72B predictions~\cite{ma2025speechllm} are included as an external reference baseline.
Our results demonstrate that shortcut reliance can be reduced in a controlled manner while preserving competitive scoring performance. Using human raters' own feature correlations as a reference point, we identify two complementary operating regimes: a \emph{human-alignment} mode that matches human reliance on proxy features, and a more aggressive \emph{malpractice-suppression} mode that further reduces shortcut dependence at an accepted cost in predictive accuracy. 
Importantly, the penalty primarily suppresses targeted proxy correlations while leaving unrelated features largely unaffected, indicating that the intervention is selective rather than causing a general degradation in model behaviour.

The main contributions of this work are:
\begin{itemize}
 \item A rank correlation penalty that controls implicit shortcut reliance in end-to-end models, operating at the output level without requiring explicit feature inputs. 
 \item Empirical evidence of shortcut reliance across both modalities, and a principled cross-modal comparison of its suppression under the proposed penalty.
 \item Two interpretable operating modes, i.e. \textit{human-alignment} and \textit{malpractice-suppression}, that can be selected by 
 assessment designer.
\end{itemize}

\section{Automated Spoken Language Assessment}
\label{sec:sla}

\begin{figure}[htbp!]
    \centering
    \includegraphics[width=0.9\linewidth]{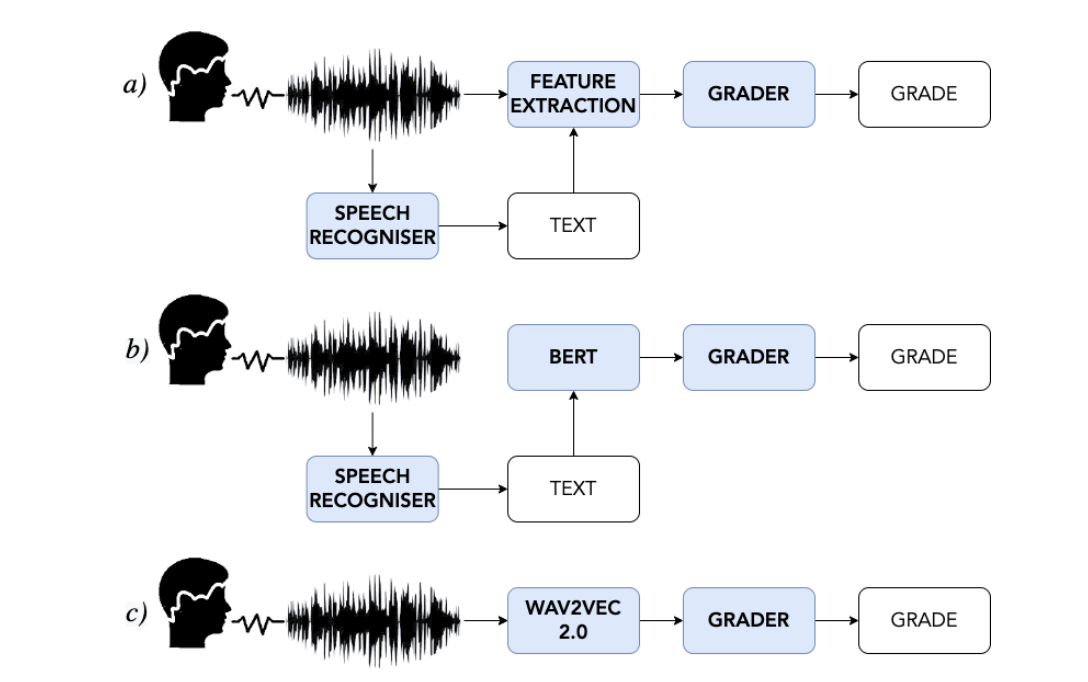}
    \caption{Three typical spoken language assessment system architectures~\cite{banno2023selfsupervised}: a) hand-crafted feature-based grader, b) text BERT-based grader, and c) audio wav2vec~2.0-based grader.}
    \label{fig:sla_automarker_architectures}
\end{figure}

Automated SLA systems assign a proficiency score $\hat{y}^{(i)}$ to a learner utterance, using the raw audio signal $x_{1:T}^{(i)}$ and/or an automatic speech recognition (ASR) transcript $w_{1:L}^{(i)}$, as input, where $T$ denotes the number of audio frames and $L$ the number of recognised word tokens. The goal is to estimate
the holistic judgement of a human examiner of this utterance $y^{(i)}$. A standard training criterion for these systems, examples of which are shown in Figure~\ref{fig:sla_automarker_architectures}, is to minimise the Mean Squared Error (MSE) between ${y^{(i)}}$ and ${\hat{y}}^{(i)}$~\cite{qian2025speak, banno2023wav2vec}
 \begin{eqnarray}
 {\cal L} = \frac{1}{n}\sum_{i=1}^n\left(y^{(i)}-{\hat y}^{(i)}\right)^2
 \end{eqnarray}
 where $n$ is the number of training examples.

\subsection{Hand-crafted feature-based auto-markers}
Early automated graders use a set of $k$ interpretable features $\hat{z}_{1:k}^{(i)}$ computed directly from the audio and transcript. A relatively simple model is then used to predict the score:
\begin{equation}
    \hat{y}^{(i)} = f_\theta\!\left(\hat{z}_{1:k}^{(i)}\right), \hat{z}_{1:k}^{(i)} \leftarrow \bigl\{x_{1:T}^{(i)},\, w_{1:L}^{(i)}\bigr\}
    \label{eq:feature-grader}
\end{equation}
SpeechRater~\cite{xi2008, zechner2009automatic, higgins2011, chen2018speechrater} and the Linguaskill Speaking auto-marker~\cite{wang2018towards, xu2021, gao2024} are examples of this form model, typical features include speech rate, pronunciation scores, measures of vocabulary and grammatical complexity.
Later iterations of SpeechRater explicitly targeted feature independence, minimising colinearity between extracted features as a design goal alongside fairness and construct relevance~\cite{chen2018speechrater}, a consideration we revisit when
selecting shortcut proxies in Section~\ref{sec:setup}.
It is important to note that each $\hat{z}_{j}^{(i)}$ is a hand-engineered proxy for a latent linguistic construct $z_{j}^{(i)}$ that the examiner implicitly takes into account. For example, the number of word tokens in an ASR transcript, $\hat{z}_j^{(i)}$, approximates the true spoken word count $z_j^{(i)}$.
While this approach is transparent and interpretable, with features directly relatable to the speaking assessment criteria, these features
are unlikely to cover all attributes of the candidate. For example higher-order aspects of proficiency such as discourse coherence and communicative achievement are very difficult to model in this way~\cite{knill2024}.


\subsection{Fine-tuned encoder-based auto-markers}
The availability of large pre-trained acoustic and language models has enabled end-to-end fine-tuning directly from the audio and transcripts with no intermediate feature extraction.
In audio-based auto-markers, self-supervised models such as wav2vec~2.0~\cite{baevski2020} are fine-tuned on the raw waveform~\cite{banno2023wav2vec, banno2023selfsupervised}, yielding:
\begin{equation}
    \hat{y}^{(i)} = f_\theta\!\left(x_{1:T}^{(i)}\right) 
    \label{eq:audio-grader}
\end{equation}


In text-based auto-markers, pre-trained foundation models such as BERT~\cite{devlin2019} and Longformer~\cite{beltagy2020longformer} are fine-tuned on the ASR transcripts~\cite{raina20_interspeech, mcknight2023automatic, qian2025speak}, giving:
\begin{equation}
    \hat{y}^{(i)} = f_\theta\!\left(w_{1:L}^{(i)}\right) 
    \label{eq:text-grader}
\end{equation}
Both types of models learn a latent representation that implicitly encodes proficiency-relevant information, rather than computing it directly and explicitly as in feature engineering. 

\subsection{Large language models as auto-markers}
More recently, large language models~(LLMs) have been evaluated as zero-shot spoken assessment systems operating on ASR transcripts~\cite{banno2025nla}.
Speech LLMs that accept raw speech with text instructions have also been examined. However the zero-shot performance of current models is not as good as the fine-tuned variants~\cite{ma2025speechllm}. Additionally, their computational cost constrains their deployment for large-scale assessment.
The present work therefore focuses on fine-tuned audio and text encoders, with zero-shot LLM predictions serving as a non-fine-tuned baseline.

\section{Shortcut Reliance in Auto-markers}
\label{sec:shortcuts}

\subsection{Measuring shortcut reliance}

Correlation between system predictions and interpretable features has long been used as a diagnostic for over-reliance in automated assessment.
In Automated Essay Scoring (AES), the early version of e-rater yielded predictions
that showed minimal differences to using essay length alone~\cite{chodorow2004beyond}.
E-rater V.2 explicitly reduced this correlation with length as a
deliberate validity criterion~\cite{attali2007construct}.
Pearson and Spearman rank correlations between essay length and model scores have since been computed as standard diagnostic quantities, and it has been shown that a model exploiting length alone can achieve state-of-the-art performance on benchmark datasets~\cite{jeon-strube-2021-countering}.

Similar concerns arise in Spoken Language Assessment (SLA).
Temporal fluency features, including speech rate, pause distribution, VAD duration, and word count, correlate with  proficiency and have been used as features in handcrafted scoring engines~\cite{wang2018towards,chen2018speechrater,bamdev2023automated}.
The validity of such measures in automated speaking assessment was examined in~\cite{handley2024measures}, observing that automatic speech evaluation systems often rely on a limited range of temporal and lexical features, and finding that articulation rate and pause frequency together account for a substantial share of variation in functional adequacy scores.
These approaches all sit within the explicit feature extraction framework previously described.

Our work addresses a distinct problem, where features are implicitly encoded by the auto-marker from  raw audio and ASR transcripts. 
Let $\hat{\mathbf{y}} = \bigl\{\hat{y}^{(1)}, \ldots, \hat{y}^{(n)}\bigr\}$
and $\mathbf{y} = \bigl\{y^{(1)}, \ldots, y^{(n)}\bigr\}$
denote model predictions and ground-truth human scores over $n$ utterances respectively.
For a given proxy, interpretable feature,  $\hat{\mathbf{z}}_j = \bigl\{\hat{z}_j^{(1)}, \ldots, \hat{z}_j^{(n)}\bigr\}$,
a shortcut is present when the model's predictions are more strongly correlated
with the proxy than the human labels are: 
\begin{equation}
    \rho\!\left(\hat{\mathbf{y}},\, \hat{\mathbf{z}}_j\right)
    \gg
    \rho\!\left(\mathbf{y},\, \hat{\mathbf{z}}_j\right),
    \label{eq:shortcut_rho}
\end{equation}
Here $\rho(\cdot,\cdot)$ denotes Spearman rank correlation (SRC) which is defined as
\begin{equation}
    \rho(\hat{\mathbf{y}},\, \mathbf{y})
    =
    \!\left(
    1
    -
    \frac{6\sum_{i=1}^{n}
          \bigl(r(\hat{y}^{(i)}) - r(y^{(i)})\bigr)^{2}}
         {n(n^2 - 1)}
    \right),
    \label{eq:src_loss}
\end{equation}
where $r(y^{(i)}) \in \{1, 2, \ldots, n\}$ denotes the rank of score $y^{(i)}$ within the batch of size $n$.
This formulation is model agnostic,
it applicable to feature-based graders, where $\hat{z}_j^{(i)}$ is an explicit model input, as well as encoder-based graders,
where $\hat{z}_j^{(i)}$ is not explicitly represented but can be computed from the raw input independently of the model.
It is also sign-agnostic, covering positive proxies, which relate to aspects of speaking where an increase correlates with an increased score (e.g.\ utterance length), and negative proxies, where the speaker wants to decrease their occurrence to get a higher score (e.g.\ disfluency count), alike.

 
\subsection{Controlling explicit shortcut reliance}

When the shortcut feature $\hat{z}_j^{(i)}$ is an explicit model input,
mitigation strategies can exploit direct access to this feature.
In prior work,
feature reweighting and the introduction of content, discourse, and lexical features reduced length dependence in early systems~\cite{chodorow2004beyond,attali2007construct}.
Content-similarity objectives were proposed to bypass length entirely by design~\cite{jeon-strube-2021-countering}, and linguistically-informed counterfactuals have been used to diagnose and penalise reliance on identified input dimensions~\cite{wang-etal-2024-beyond-agreement}.

Another class of approaches uses post-hoc interpretability tools to diagnose shortcut reliance.
Integrated gradients (IG)~\cite{sundararajan2017axiomatic} attribute an importance score to each input dimension by integrating the model's gradient along the path from a neutral baseline to the actual input, and have been applied to reveal that AES models exploits shallow surface cues, such as specific trigger words, in the input to drive model predictions~\cite{kumar-etal-2023-automatic}. 
In~\cite{kumar-etal-2023-automatic} IG remain diagnostic tools rather than training-time penalties.
The Right for the Right Reasons~(RRR) framework~\cite{ross2017right} extends this explainability into the training process by penalising the gradient of the prediction with respect to shortcut features in addition to the general task-specific loss $\mathcal{L}_{\mathrm{task}}$, in a simplified form:
\begin{equation}
    \mathcal{L}
    =
    \mathcal{L}_{\mathrm{task}}
    +
    \frac{\lambda}{n}
    \sum_{i=1}^{n}
    \left(
        \frac{\partial \hat{y}^{(i)}}{\partial \hat{z}_j^{(i)}}
    \right)^{\!2}.
    \label{eq:rrr}
\end{equation}
Right for Better Reasons~\cite{shao2021right} further extends this framework by replacing input gradients with influence functions as the constraint signal, demonstrating more effective correction of shortcut behaviour.

A related line of work corrects shortcut behaviour directly at the representation level:
COrrect and MItigate (COMI)~\cite{zhao2024comi} introduces a shortcut margin loss that adaptively suppresses the feature weights of identified shortcuts, using attribution-based methods such as Local Interpretable Model-agnostic Explanations (LIME) to locate shortcut tokens in text inputs.
Focus-and-Ignore Low-Rank Adaptation (FiLoRA)~\cite{chung2026filora} decomposes model adaptation into feature-group-aligned LoRA modules and applies instruction-conditioned gating, allowing a natural language instruction to selectively amplify or suppress identified core or spurious feature groups in multimodal models.

However, these approaches share a common requirement: the shortcut feature must be an explicit, accessible input to the model, so that it can be removed, reweighted, or targeted by a gradient-based penalty.
Fine-tuned encoder-based auto-markers do not satisfy these conditions.
A ModernBERT grader operating on ASR transcripts, or a wav2vec~2.0 grader operating on raw waveforms, does not receive word count or speech duration as an input. Instead, length information is absorbed implicitly into learned representations via attention over token sequences and position encodings.
That is, for encoder-based graders there is no direct gradient path from $\hat{y}^{(i)}$ to $\hat{z}_j^{(i)}$; Equation~\ref{eq:rrr} is therefore inapplicable.

While the feature-level solution cannot transfer to encoder-based graders, the correlation-based diagnostic in Equation~\ref{eq:shortcut_rho} can:
$\hat{z}_j^{(i)}$, such as word count, speaking rate, or silence duration, can be computed from the raw input independently of the model and compared against $\hat{y}^{(i)}$ at the output level.
The following section introduces a training-time penalty that exploits this diagnostic without requiring explicit feature access. It is applicable to both feature-based and encoder-based graders within a unified framework.


\section{Controlling Implicit Shortcut Reliance}
\label{sec:implicit}

This section proposes a method to control the reliance on implicit shortcuts. As previously noted, the aim is to break the over-reliance of the auto-marker on one or more interpretable feature whilst maintaining the same relationship that the feature(s) has to human judgement.
The proposed method is applicable to any 
fine-tuned encoder-based speech or text grader and can penalise reliance on any computable feature. As these
features must by definition be 
human interpretable, it is also useful for safeguarding against known test malpractice strategies.
Any feature that can be communicated as actionable advice (e.g.\ ``speak for longer'' or ``throw in more adjectives'') is by definition interpretable and computable so can be targeted by this approach.

Instead of the standard MSE loss as mentioned in Section~\ref{sec:sla}, we adopted a rank-based loss because the dynamic ranges of holistic scores and individual proxy features differ substantially:
for example, in the Speak~\&~Improve 2025 Corpus~\cite{sicorpus25,qian2025speak} holistic scores lie in the range $[1, 6]$ whilst word counts can exceed $100$.
Following~\cite{fathullah2024needs}, it is possible to use the SRC as the task training criterion.
However since the rank operator $r(\cdot)$ is discrete and non-differentiable, it cannot be directly optimised via gradient descent. Both $r(\hat{y}^{(i)})$ and $r(y^{(i)})$ are replaced by the differentiable approximation of~\cite{blondel2020fast}, yielding a fully differentiable objective.
In the case of tied values, averaged ranks are assigned:
for example, if two samples share rank~2, both receive rank~2.5, which is the standard convention for Spearman correlation, and is handled automatically by this open-source implementation\footnote{\url{https://github.com/google-research/fast-soft-sort}}.

We propose to add one or more regularisation terms to Equation~\ref{eq:src_loss}. 
Let $\mathcal{J} \subseteq \{1, \ldots, k\}$ denote the set of proxy indices
selected for regularisation.
For each $j \in \mathcal{J}$,
let $\hat{\mathbf{z}}_j = \bigl\{\hat{z}_j^{(1)}, \ldots, \hat{z}_j^{(n)}\bigr\}$
denote the values of proxy $j$ across $n$ utterances.
The full training objective to minimise is:
\begin{equation}
    \mathcal{L}
    =
    -\rho\!\left(\hat{\mathbf{y}},\, \mathbf{y}\right)
    +
    \sum_{j \in \mathcal{J}}\,
    \lambda_j\,
    \rho\!\left(\hat{\mathbf{y}},\, \hat{\mathbf{z}}_j\right),
    \label{eq:full_loss}
\end{equation}
where $\mathbf{y}$ and ${\hat{\mathbf{y}}}$ are the reference and prediction scores respectively and $\lambda_j \geq 0$ controls the strength of penalisation for proxy $j$. Thus
minimising $\mathcal{L}$ simultaneously maximises prediction--score correlation
and minimises prediction--proxy correlation.
The choice of $\lambda_j$ determines the level to which the correlation with the interpretable feature, proxy, $\hat{\mathbf{z}}_j$ is supressed. It can be used to reduce the correlation to the level seen in the reference scores, $\rho(\mathbf{y},{\hat{\mathbf{z}}_j})$, or set to higher values to penalise candidates who attempt to use this form of shortcut.

\section{Experimental set-up}
\label{sec:setup}

\subsection{S\&I data}

Experiments are conducted on the Speak~\&~Improve~(S\&I) corpus~\cite{sicorpus25,  qian2025speak},
a large-scale dataset of spoken English responses from L2 learners collected through the Speak~\&~Improve online practice platform\footnote{\url{https://speakandimprove.com}}.
Responses span four open speaking task types:
Part~1 (interview: short responses to personal questions),
Part~3 (long turn: giving an opinion on a topic),
Part~4 (long turn: describing a diagram),
and Part~5 (communication activity: responding to questions on a topic).
Each part is assigned a holistic proficiency score $y^{(i)} \in [1, 6]$ 
corresponding approximately to CEFR levels A1 to C1+.
The corpus is divided into training (6{,}642 submissions, 39{,}490 utterances), development (438 submissions, 5{,}616 utterances), and evaluation (300 submissions, 3{,}209 utterances) sets~\cite{qian2025speak}.

Table~\ref{tab:combined_feature_corr_eval} reports Spearman rank correlations between four candidate features and the holistic reference score, alongside the feature--feature correlations that motivate our choice of shortcut proxy for each grader. The candidate text-based features selected are the a) number of words (\#words) and b) number of unique words (\#uniq), both computed from the ASR transcript. For audio-based features, a) average ASR confidence (ASR conf) per word and b) the candidate's speaking time in seconds as measured by a voice activity detection system (VAD time), were selected for this paper.

\begin{table}[ht]
\begin{center}
\caption{Spearman $\rho$ between text- and audio-based features, and holistic reference score for the S\&I evaluation set.}
\begin{tabular}{l|c|cccc}
\toprule
\textbf{Features} & \textbf{Ref score} & \textbf{\#words} & \textbf{\#uniq.} & \textbf{ASR conf} & \textbf{VAD time} \\
\midrule
{\#words}       & 0.659 & 1.000 & 0.953 & 0.446 & 0.655 \\
{\#uniq.} & 0.680 & --    & 1.000 & 0.486 & 0.624    \\
{ASR conf}  & 0.657 & --    & --    & 1.000 & 0.234 \\
{VAD time}    & 0.450 & --    & --    & --    & 1.000 \\
\bottomrule
\end{tabular}
\label{tab:combined_feature_corr_eval}
\end{center}
\end{table}

For the text-based grader, 
we adopt word count $\hat{z}_{\mathrm{wc}}^{(i)}$, derived directly from the \texttt{Whisper-small.en}\footnote{https://huggingface.co/openai/whisper-small.en} transcript $w_{1:L}^{(i)}$, as the shortcut proxy $\hat{\mathbf{z}}_j$, since it is the more directly actionable quantity (a learner can be advised to ``speak for longer'', whereas ``use more unique words'' is a less concrete strategy).
Unique word count is used as an internal consistency check:
we expect the penalty applied to word count to suppress reliance on unique word count too given their high correlation. 
Average ASR confidence per word is 
used as a selectivity check:
an effective penalty should leave its correlation with model predictions largely intact, providing evidence that the intervention is targeted rather than a general degradation.

For the audio-based grader, we examine  VAD time $\hat{z}_{\mathrm{VAD}}^{(i)}$, 
computed directly from the waveform $x_{1:T}^{(i)}$ using the Praat VAD algorithm, as the shortcut proxy, since it reflects the fact that longer spoken responses tend to contain more speech content.
Word count and average ASR confidence per word are also used consistency checks to see the impact on other attributes. 


\subsection{Models}

All graders are trained at the response level: one example per response, sharing the part-level holistic label as the target. For the two fine-tuned graders, the shortcut regularisation strength $\lambda_j$ is swept over $[0, 0.30]$, and the differentiable rank approximation of~\cite{blondel2020fast} (Section~\ref{sec:implicit}) uses a default regularisation strength of $0.1$ throughout.

\subsubsection{ModernBERT and Qwen2.5-72B}
Both text-based graders use the ASR transcript $w_{1:L}^{(i)}$ from \texttt{Whisper-small.en}.
For the ModernBERT-based grader, the architecture follows the BERT-based grader of~\cite{qian2025speak} apart from the encoder being \texttt{ModernBERT-base}\footnote{https://huggingface.co/answerdotai/ModernBERT-base}: token-level embeddings from the encoder are pooled into an utterance-level representation
using four parallel self-attention heads,
and passed through a two-layer feedforward network with ReLU activations to a scalar output $\hat{y}^{(i)} = f_\theta(w_{1:L}^{(i)})$.
For each $\lambda_\mathrm{WC}$, ten models are trained with different random seeds and predictions are averaged across the ensemble. The training objective follows Equation~\ref{eq:full_loss}, with word count $\hat{\mathbf{z}}_\mathrm{wc}$ as the shortcut proxy ($|\mathcal{J}| = 1$). Zero-shot predictions from Qwen2.5-72B~\cite{qwen2025qwen25}, following the natural language-based assessment~approach~\cite{banno2025nla}, as a system that is not fine-tuned.

\subsubsection{wav2vec~2.0}
The audio-based grader uses wav2vec~2.0-base~\cite{baevski2020} as the pre-trained encoder, with the standard mean-pooling output layer replaced by four attention heads that pool frame-level representations into a single utterance vector, consistent
with~\cite{mcknight2023automatic}.
The pooled representation is passed through a two-layer feedforward head (hidden size $4 \times 768$, ReLU with dropout before and after) to a scalar output, $\hat{y}^{(i)} = f_\theta(x_{1:T}^{(i)})$.
The model is fine-tuned with AdamW. 
Parts 3 and 4 are trained with batch size~16, while Parts 1 and 5 are trained with batch size~64 to increase within-batch rank variablility (since they contain multiple short responses sharing one part-level label).
All graders are trained for 2 epochs and no seed ensembling is used, owing to the higher cost of training on raw waveforms.
The training objective follows Equation~\ref{eq:full_loss}, with VAD time $\hat{\mathbf{z}}_\mathrm{VAD}$ as the shortcut proxy ($|\mathcal{J}| = 1$).

\subsection{Score Calibration and Combination}

Part-level predictions from models trained with different $\lambda$ values are combined into a single submission-level score for computing Spearman rank correlation across the full test set. 
As the SRC loss function does not impose any constraints on the absolute values of any predictions, only the ranking, some form of score calibration is required.
A standard approach is to apply linear calibration of part predictions to the reference score scale~\cite{qian2025speak}.
This was considered but rejected here: at high $\lambda$, the penalty can drive
$\rho(\hat{\mathbf{y}}, \mathbf{y})$ sufficiently low that the calibration slope $a$ in
$\tilde{y}_\mathrm{cal}^{(p,i)} = a\,\hat{y}^{(p,i)} + b$ 
becomes negative.
Since
$\rho(\tilde{\mathbf{y}}_\mathrm{cal}, \hat{\mathbf{z}}_j)
= a \cdot \rho(\hat{\mathbf{y}}, \hat{\mathbf{z}}_j)$,
a negative slope flips the sign of the proxy correlation, causing a spurious bounce-back in the measured $\lambda$ sweep.
Henceforth, Z-score normalisation is adopted, since it operates purely on the prediction distribution without reference to $\mathbf{y}$ and therefore cannot induce a sign flip.

For each part $p \in P = \{\text{P1, P3, P4, P5}\}$ and each value of $\lambda_j$, the part-level predictions
$\hat{\mathbf{y}}^{(p)} = \bigl\{\hat{y}^{(p,i)}\bigr\}_{i=1}^{n_p}$
are normalised as:
\begin{equation}
    \tilde{y}^{(p,i)}
    =
    \frac{\hat{y}^{(p,i)} - \mu^{(p)}}{\sigma^{(p)}},
    \label{eq:zscore}
\end{equation}
where $\mu^{(p)}$ and $\sigma^{(p)}$ are the mean and standard deviation of predictions for part $p$, and $n_p$ is the number of utterances in part $p$.
The submission-level score for utterance $i$ is then the average of normalised part predictions:
\begin{equation}
    \tilde{y}^{(i)}
    =
    \frac{1}{|P|}
    \sum_{p \in P}
    \tilde{y}^{(p,i)},
    \label{eq:combination}
\end{equation}
where each part contributes equally regardless of prediction scale.
Per-part Spearman rank correlations are computed directly from $\hat{\mathbf{y}}^{(p)}$ before normalisation. Z-score normalisation is applied only for submission-level
combination.


\section{Experiment results}

\subsection{Text as input: ModernBERT}
\label{sec:modernbert_results}

Table~\ref{tab:score_feature_corr_eval} reports Spearman correlations between ModernBERT predictions and the reference score and three text-derived features, alongside the corresponding human reference correlations.
Using SRC with no additional regularisation rather than MSE loss has negligible effect on both the baseline performance
and level of correlation with features.
Both ModernBERT-based models exhibit substantially higher correlations with word count and unique word count
than the human reference exhibits with the same features. This
illustrates an over-reliance, shortcut, on these features based on  Equation~\ref{eq:shortcut_rho}.
Conversely correlations with average ASR confidence are consistent with the human reference.
The zero-shot Qwen2.5-72B feature correlations are more consistent with the reference correlations. It does not have the 
very high correlations of the ModernBERT system with the number of words and unique words. This is felt to be because
the zero-shot model has had no fine-tuning to the task, thus it has no opportunity to learn task-specific shortcuts, 


\begin{table}[ht!]
\caption{Baseline Text models: score and feature correlations in the S\&I evaluation set.}
\begin{center}
\begin{tabular}{llccccc}
\toprule
\textbf{Model} & \textbf{Loss} & \textbf{Ref} & \textbf{\#words} & \textbf{\#uniq.} & \textbf{ASR conf} \\
\midrule
Reference & -- & 1.000 & 0.659 & 0.680 & 0.657 \\
\midrule
\multirow{2}{*}{ModernBERT}
 & MSE & 0.758 & 0.872 & 0.911 & 0.653 \\
 & SRC & 0.761 & 0.874 & 0.911 & 0.655 \\
\midrule
Qwen2.5-72B & -- & 0.755 & 0.733 & 0.792 & 0.719 \\
\bottomrule
\end{tabular}
\label{tab:score_feature_corr_eval}
\end{center}
\end{table}

\begin{figure}[htbp!]
    \centering
    \includegraphics[width=0.88\linewidth, trim={0.2cm 0cm 20.23cm 0cm}, clip]{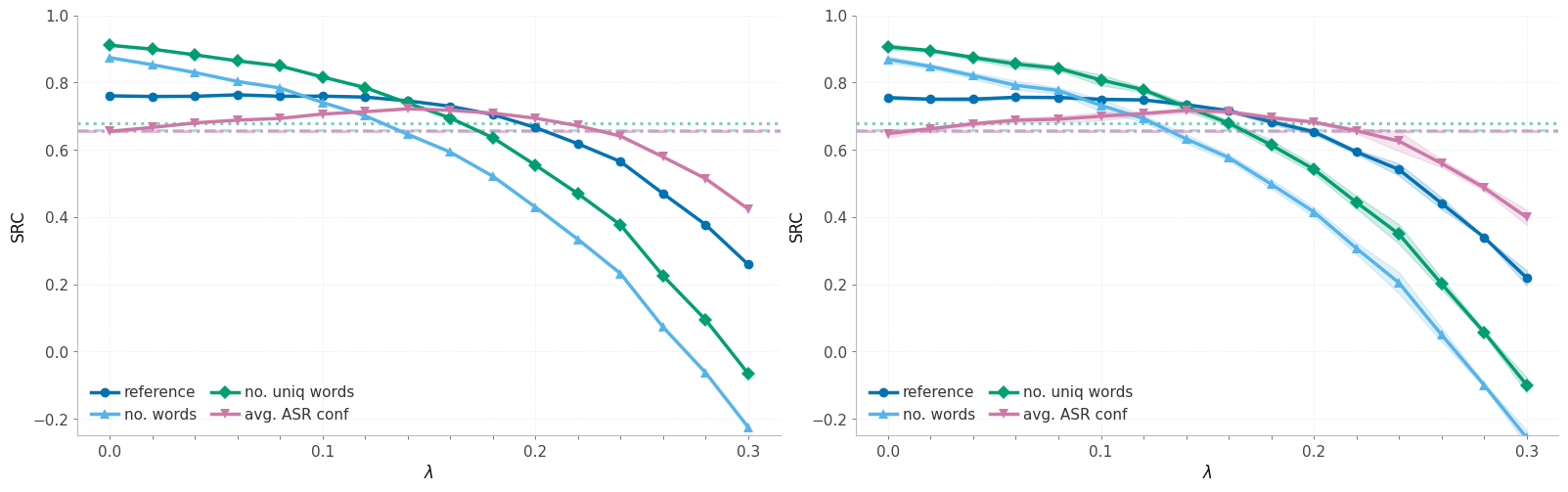}
    \caption{Spearman $\rho$ between ModernBERT-based models' predictions and reference features across $\lambda$ values. Dashed horizontal lines indicate the Spearman $\rho$ between reference scores and each feature. (S\&I evaluation set)}
    \label{fig:modernbert_sweep}
\end{figure}

To reduce the influence of the number of words~\footnote{Word-count rather than the number of unique words was used as this is less sensitive to the length of the response.} on the ModernBERT score a regularisation term targeting this feature was introduced 
wih a controllable weight $\lambda_\mathrm{WC}$.
Figure~\ref{fig:modernbert_sweep} shows how feature and score  correlations evolve as $\lambda_\mathrm{WC}$ is swept over $[0, 0.30]$.
As $\lambda_\mathrm{WC}$ increases, correlation with word count decreases steadily, crossing the human reference level at $\lambda_\mathrm{WC} \approx 0.13$, which defines the \textit{human-alignment} operating mode, where shortcut reliance on word count matches the level exhibited by human raters 
($\rho(\hat{\mathbf{y}}, \hat{\mathbf{z}}_\mathrm{WC})
\approx \rho(\mathbf{y}, \hat{\mathbf{z}}_\mathrm{WC})$).
Accuracy, $\rho(\hat{\mathbf{y}}, \mathbf{y})$, remains at a competitive level.
Beyond this $\lambda_\mathrm{WC}$ point, both accuracy and word-count correlation continue to decrease more sharply, defining the \textit{malpractice-suppression} mode and reflecting the cost of shortcut suppression.
The unique word count shows a similar pattern to word count, confirming that the penalty has a consistent effect on highly correlated shortcut features.
Features that are not shortcut proxies, here represented by average ASR confidence, are minimally affected by the penalty until broader degradation sets in at high $\lambda_\mathrm{WC}$, providing evidence that the intervention is targeted rather than a general suppression of model reliance.

\subsection{Audio as input: wav2vec~2.0}
\label{sec:wav2vec2_results}

Table~\ref{tab:wav2vec_score_feature_corr_eval} shows that wav2vec~2.0 exhibits higher correlation with VAD time and word count than the human reference, confirming over-reliance consistent with Equation~\ref{eq:shortcut_rho}.
As with the text-based grader, average ASR confidence shows no over-reliance at baseline.

\begin{figure}[h!]
    \centering
    \includegraphics[width=0.9\linewidth]{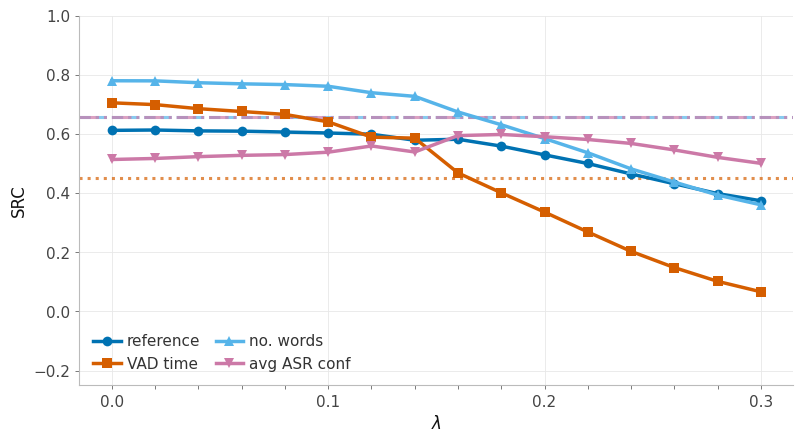}
    \caption{Spearman $\rho$ between wav2vec-based model predictions and reference features across $\lambda$ values. Dashed horizontal lines indicate the Spearman $\rho$ between reference scores and each feature. (S\&I evaluation set)}
    \label{fig:wav2vec_sweep}
\end{figure}

\begin{table}[ht!]
\caption{Audio models: score and feature correlations in the S\&I evaluation set.}
\begin{center}
\begin{tabular}{lcccc}
\toprule
\textbf{Model} & \textbf{Ref score} & \textbf{VAD time} & \textbf{\#words} & \textbf {ASR conf} \\
\midrule
 Reference        & 1.000 & 0.450 & 0.659 & 0.657 \\
 wav2vec~2.0         & 0.612 & 0.705 & 0.780 & 0.514 \\ 
\bottomrule
\end{tabular}
\label{tab:wav2vec_score_feature_corr_eval}
\end{center}
\end{table}
Again a regularisation term targeting in this case the VAD time with weight $\lambda_\mathrm{VAD}$ was introduced.
Figure~\ref{fig:wav2vec_sweep} shows how these correlations evolve as $\lambda_\mathrm{VAD}$ is swept over $[0, 0.30]$.
Correlation with VAD time decreases steadily as $\lambda_\mathrm{VAD}$ increases, and crosses the human reference level at $\lambda_\mathrm{VAD} \approx 0.17$
($\rho(\hat{\mathbf{y}}, \hat{\mathbf{z}}_\mathrm{VAD})
\approx \rho(\mathbf{y}, \hat{\mathbf{z}}_\mathrm{VAD})$),
marking the \textit{human-alignment} operating mode at a moderate cost to accuracy, $\rho(\hat{\mathbf{y}}, \mathbf{y})$.
Past this point, both continue to decline more sharply, entering the \textit{malpractice-suppression} mode.
A similar trend is seen for word count, consistent with its relatively high correlation with VAD time (Table~\ref{tab:combined_feature_corr_eval}).
Average ASR confidence, representing a non-shortcut feature, remains largely stable until high $\lambda_\mathrm{VAD}$, again pointing to a targeted rather than general suppression of model reliance, which is consistent with what is observed in the text-based grader (Section~\ref{sec:modernbert_results}).

\section{Conclusions and Future Work}

This paper introduced a rank correlation penalty that discourages fine-tuned encoder-based graders from over-relying on externally computable proxy features, operating at the output level without requiring access to or modification of the encoder's internal representations.
Unlike prior explicit feature-level mitigation strategies, which presuppose a direct gradient path from the predicted score to the shortcut feature, the proposed penalty applies to both text-based and audio-based graders within a unified framework.
Experiments on the Speak~\&~Improve corpus confirmed shortcut reliance in both a ModernBERT text-based grader and a wav2vec~2.0 audio-based grader, and demonstrated that the penalty reduces this over-reliance in a controllable manner, yielding two natural operating modes: a \textit{human-alignment} mode, where shortcut reliance is reduced to match the human reference level with minimal cost to accuracy,;
and a \textit{malpractice-suppression} mode, where reliance is reduced further at an accepted cost to accuracy.
Across both modalities, features that are highly correlated with the target shortcut proxy are affected by the penalty in a consistent way; features that are not shortcut proxies are minimally affected by the penalty until broader degradation sets in, providing evidence that the intervention is targeted rather than a general suppression of model reliance.

Several directions remain open.
The zero-shot Qwen2.5-72B baseline exhibits some verbosity bias, suggesting shortcut reliance is not entirely absent from large pre-trained models; extending the approach to fine-tuned LLM-based graders is a natural extension.
The general formulation in Equation~\ref{eq:full_loss} supports simultaneous penalisation over a set $\mathcal{J}$ of proxies, enabling multiple exploitable features to be targeted at once.
Finally, all experiments used holistic monologic assessment; extending the approach to dialogic assessment
%
is another natural next step.

\section*{Acknowledgment}
This paper reports on research supported by Cambridge University Press \& Assessment, a department of The Chancellor, Masters, and Scholars of the University of Cambridge.
The authors would like to thank the ALTA Spoken Language Processing Technology Project Team for general discussions and contributions to the evaluation infrastructure.

No use was made of Generative AI in the writing of this paper.



\bibliographystyle{IEEEtran}
\bibliography{references}

\end{document}